\documentclass{article}
\usepackage{amssymb}
\usepackage{xcolor}
\usepackage{graphicx}
\usepackage[implicit=false]{hyperref}

\long\def\comment#1{}

\newtheorem{assumption}{Assumption} 
\newtheorem{theorem}{Theorem}

\newtheorem{proposition}[theorem]{Proposition}

\newenvironment{oldtheorem}[1]{
 \bigskip \noindent
 {\bf Theorem~\ref{#1}} \em}{\bigskip}
 
\newcommand{\Thm}[1]{Theorem~\ref{#1}}

\newcommand{\Ass}[1]{Assumption~\ref{#1}} 
\newcommand{\Fig}[1]{Figure~\ref{#1}} 
\newcommand{\Tab}[1]{Table~\ref{#1}} 
\newcommand{\Eq}[1]{Equation~\ref{#1}} 
\newcommand{\Eqs}[1]{Equations~\ref{#1}} 
\newcommand{\qed}{\mbox{$\Box$}} 
\newcommand{\Sec}[1]{Section~\ref{#1}} 
 
\newcommand{\Bs}{B_S} 
\newcommand{\Bp}{B_P} 
\newcommand{\Bsc}{B_{S_C}} 
\newcommand{\Bscp}{B_{S'_C}} 
\newcommand{\Bsone}{B_{S1}} 
\newcommand{\Bstwo}{B_{S2}}

\newcommand{\eBs}{B^e_S}

\newcommand{\eBsc}{B^e_{S_C}} 
\newcommand{\eBscp}{B^e_{S'_C}} 
\newcommand{\eBsone}{B^e_{S1}} 
\newcommand{\eBstwo}{B^e_{S2}}

\newcommand{\Csone}{C_{S1}} 
\newcommand{\Cstwo}{C_{S2}}

\newcommand{\eCs}{C^e_S}

\newcommand{\eCsone}{C^e_{S1}} 
\newcommand{\eCstwo}{C^e_{S2}}

\newcommand{\dXdY}[2]{\partial {#1} / \partial {#2}} 
 
\newcommand{\BG}{BG} 
\newcommand{\BGe}{BGe} 
\newcommand{\BGp}{BGp}

\newcommand{\m}{\vec{m}} 
\newcommand{\vecv}{\vec{v}} 
\newcommand{\vecb}{\vec{b}} 
\newcommand{\vecbi}{\vec{b_i}} 
 
\newcommand{\matW}{W} 
\newcommand{\matS}{S} 
\newcommand{\matSigma}{\Sigma} 
\newcommand{\matTau}{T} 
\newcommand{\x}{\vec{x}} 
\newcommand{\xone}{\vec{x_1}} 
\newcommand{\xl}{\vec{x_l}} 
\newcommand{\Xl}{\overline{X}_l} 
\newcommand{\vecx}{\vec{x}} 
\newcommand{\vecxi}{\vec{x_i}} 
\newcommand{\vecmu}{\vec{\mu}} 
 
\newcommand{\vecbij}{b_{ij}}

\newcommand{\matB}{B} 
\newcommand{\tr}{\mbox{\it tr}}

\newcommand{\Dil}{D^{x_i \Pi_i}_l} 
\newcommand{\DiPi}{D^{\Pi_i}_l} 
\newcommand{\Di}{D^{x_i \Pi_i}} 
\newcommand{\vecmuD}{\vec{\mu}_D} 
\newcommand{\Bsdp}{B_{S'}} 
\newcommand{\eBsdp}{B^e_{S'}}

\title{Learning Gaussian Networks} 

\comment{
\author{ 
Dan Geiger\thanks{Author's primary affiliation: Computer 
Science Department, Technion, Haifa 32000, Israel.}\ \ \ \   
David Heckerman\\ 
\\ 
Microsoft Research, Bldg 9S\\ 
Redmond WA 98052-6399\\ 
dang@cs.technion.ac.il, heckerma@microsoft.com} 
}

\author{
Dan Geiger\\ 
geiger02@gmail.com\\
\\
David Heckerman\\
heckerma@hotmail.com}

\date{July 1994, Revised May 2021}

\begin{document}

\maketitle
  
\begin{abstract} 
 
\noindent We describe scoring metrics for learning Bayesian networks from a combination of user knowledge and statistical data.  Previous work has concentrated on metrics for domains containing only discrete
variables, under the assumption that data represents a multinomial
sample.  In this paper, we extend this work, developing 
scoring metrics for domains containing only continuous variables
under the assumption that continuous data is sampled from
a multivariate normal distribution.  Our work extends traditional
statistical approaches for identifying vanishing regression
coefficients in that we identify two important assumptions,
called {\em event equivalence} and {\em parameter modularity},
that when combined allow the construction of prior
distributions for multivariate normal parameters from a single
{\em prior Bayesian network} specified by a user.  
 
\bigskip

\noindent
Corrections to the original text in \textcolor{red}{red} are taken
from the 2021 update of 
J. Kuipers, G. Moffa, and D. Heckerman, Addendum on the scoring
of Gaussian directed acyclic graphical models.
{\em Annals of Statistics} 42, 1689-1691, Aug 2014
(\href{https://arxiv.org/abs/1402.6863}{\underline{arXiv:1402.6863}}).
Other updates to the original are in \textcolor{blue}{blue}.

\end{abstract}

\section{Introduction} 

Several researchers have examined methods for learning Bayesian networks from data, including Cooper and Herskovits (1991,1992)\nocite{Cooper91,Cooper92}, Buntine (1991), Spiegelhalter et al. (1993)\nocite{SDLC93},
and Heckerman et al. (1994) (herein referred to as CH, Buntine, SDLC,
and HGC, respectively).  These methods all have the same basic components: a scoring metric and a search procedure.
The metric computes a score that is proportional to the posterior
probability of a network structure, given data and a user's prior
knowledge.  The search procedure generates networks for evaluation by the scoring metric.  These methods use the two components to 
identify a network or set of networks with high relative posterior probabilities, and these networks are then used to predict future events.  

Previous work has concentrated on domains containing only discrete
variables, under the assumption that data is sampled
from a multivariate discrete distribution.
In this paper, we develop metrics for
domains containing only continuous variables, 
under the assumption that continuous data is sampled from
a multivariate normal (Gaussian) distribution.  
Previously, when working with
continuous variables, the standard solution had been to transform each such variable $x_i$ to a discrete one by splitting its domain into several mutually exclusive and exhaustive regions.  Our metrics eliminate
the need for this transformation.  In addition,
our metrics have the advantage that they use the low polynomial dimentionality of the parameter space of a mulitivariate normal distribution, whereas their discrete counterparts often require a parameter space that is exponential in the number of domain variables.

Our work can be viewed as an extension of traditional
statistical approaches for identifying vanishing regression
coefficients, such as those described in DeGroot (1970, Chapter 11)\nocite{DeGroot70}.
In particular, we translate two assumptions
that we identified in HGC for domains containing only discrete
variables, called parameter modularity and event equivalence, to 
domains containing continuous variables.  
The assumption of {\em parameter modularity,} addresses the 
relationship among prior distributions of parameters for different 
Bayesian-network structures.  The property of {\em event equivalence} says that two Bayesian-network structures that represent the same set of independence assertions should
correspond to the same event and thus receive the same score.  
We show that, when combined, these assumptions allow the construction of reasonable prior distributions for multivariate normal parameters 
from a single
{\em prior Bayesian network} specified by a user.  

Our identification
of event equivalence arises from a 
subtle distinction between two types of Bayesian networks.  The first 
type, called {\em belief networks}, represents only assertions of 
conditional independence and dependence.  The second type, called {\em causal 
networks}, represents assertions of cause and effect as well as 
assertions of independence and dependence.  In this paper, we argue that 
metrics for belief networks should satisfy event equivalence, whereas
metrics for causal networks need not.
 
Our score-equivalent metrics for belief networks are similar to the metrics described by Dawid and Lauritzen (1993)\nocite{DL93}, except that our metrics score directed networks, whereas their metrics score undirected networks.  In this paper, we concentrate on directed models rather than on undirected models, because we believe that users find the former easier to build and interpret.  

We note that much of the mathematics involved in our derivations is 
borrowed from DeGroot's book, ``Optimal Statistical Decisions,'' (1970).

\section{Gaussian Belief Networks} \label{sec:gn}

Throughout this discussion, we consider a domain $\x$ of $n$ 
continuous variables $x_1,\ldots,x_n$.  We use $\rho(\x|\xi)$
to denote the joint probability density function (pdf) over $\x$
of a person with background knowledge $\xi$.
We use $p(e|\xi)$ to denote the probability of a discrete event $e$. 

A belief network for $\x$ represents a joint pdf over $\x$ by encoding assertions of conditional independence as well as a collection of pdfs.  From the chain rule of
probability, we know
\begin{equation} \label{eq:chain}
\rho(x_1,\ldots,x_n|\xi) = 
  \prod_{i=1}^n \rho(x_i|x_1,\ldots,x_{i-1},\xi)
\end{equation}
For each variable $x_i$, 
let $\Pi_i \subseteq \{x_1,\ldots,x_{i-1}\}$ be a
set of variables that renders $x_i$ and $\{x_1,\ldots,x_{i-1}\}$ conditionally independent.  That is,
\begin{equation} \label{eq:i-pi}
\rho(x_i|x_1,\ldots,x_{i-1},\xi) = \rho(x_i|\Pi_i,\xi)
\end{equation}
A belief network is a pair $(\Bs,\Bp)$, 
where $\Bs$ is a belief-network structure
that encodes the assertions of 
conditional independence in
\Eq{eq:i-pi}, and $\Bp$ is a set of pdfs corresponding to that structure.  
In particular, $\Bs$ is a directed acyclic graph such that
(1) each variable in $U$ corresponds to a node in $\Bs$, and (2) the
parents of the node 
corresponding to $x_i$ are the nodes corresponding to
the variables in $\Pi_i$.  (In the remainder of this paper, we use $x_i$
to refer to both the variable and its corresponding node in a graph.)  
Associated with node $x_i$ in $\Bs$ are the pdfs $\rho(x_i|\Pi_i,\xi)$.  $\Bp$ is the union of these
pdfs.  Combining Equations~\ref{eq:chain} and \ref{eq:i-pi}, we
see that any belief network for $\x$ uniquely determines a joint pdf for $\x$.  That is,
\begin{displaymath} \label{eq:prod-decomp}
\rho(x_1,\ldots,x_n|\xi) = \prod_{i=1}^n \rho(x_i|\Pi_i,\xi)
\end{displaymath}
A {\em minimal belief network} is a belief network where
\Eq{eq:i-pi} is violated if any arc is removed.
Thus, a minimal belief network represents
both assertions of independence and assertions of dependence.

Let us suppose that  
the joint probability density function for $\x$ is a multivariate (nonsingular) normal distribution.  In this case, we write
\begin{eqnarray*}  \label{eq:normal}      
\lefteqn{\rho(\x|\xi) = n(\m,\matSigma^{-1}) } \\
& \equiv &
(2 \pi)^{-n/2} |\matSigma|^{-1/2} 
e^{-1/2 (\x - \m)' \matSigma ^{-1} (\x - \m)} 
\end{eqnarray*}
where $\m$ is an $n$-dimensional mean  
vector, and $\matSigma = (\sigma_{ij})$ is an 
$n \times n$ covariance  
matrix, both of which are implicitly functions of $\xi$, and
where 
$|\matSigma|$ is the determinant of 
$\matSigma$. 
We shall often find it convenient  
to refer to the {\em precision matrix}  
$\matW = \matSigma^{-1}$, whose elements  
are denoted by $w_{ij}$. 
 
This distribution
can be written as a product 
of conditional distributions 
each being an independent normal distribution. Namely, 
\begin{equation} \label{eq:prod-n} 
\rho(\x|\xi)= \prod_{i=1}^n 
\rho(x_i | x_1, \ldots,x_{i-1},\xi) 
\end{equation} 
\begin{equation} 
\label{eq:cond-normal} 
\rho(x_i | x_1, \ldots,x_{i-1},\xi) =  
   n(m_i + \sum_{j=1}^{i-1} \vecbij (x_j - m_j), 1/v_i) 
\end{equation} 
where $m_i$ is the unconditional mean of $x_i$, 
$v_i$ is the conditional variance of $x_i$ given   
values for $x_1, \ldots, x_{i-1}$, and
$b_{ij}$ is a linear coefficient reflecting the
strength of the relationship between $x_i$ and $x_j$
(e.g., DeGroot, p.55).\footnote{The coefficients $b_{ij}$ can
be thought of as regression coefficients or expressed
in terms of Yule's (1907)\nocite{Yule07} partial regression coefficient
$\beta$.}  Thus, we may interpret a multivariate normal distribution
as a belief network, where $b_{ij} = 0$ ($j<i$) implies that
$x_j$ is not a parent of $x_i$.  We call this special form of a belief network a Gaussian belief network.
The name is adopted from Shachter and Kenley (1989) 
\nocite{Shachter89b} who first described 
Gaussian influence diagrams. 

More formally, a {\em Gaussian belief network} is 
a pair $(\Bs,\Bp)$, where (1) $\Bs$ is a belief-network structure containing nodes $x_1,\ldots,x_n$ and no arc from 
$x_j$ to $x_i$ whenever $b_{ij} = 0, j<i$, (2) $\Bp$ is the collection of parameters $\vec{m} = (m_1,\ldots,m_n)$, $\vecv = \{v_1, \ldots, v_n\}$, and $\{b_{ij} \mid j < i \}$, and (3) the joint distribution over $\x$ is determined by \Eqs{eq:prod-n} and \ref{eq:cond-normal}.
Due to special properties of nonsingular normal distributions, a {\em minimal
Gaussian belief network} is one were there is an arc from 
$x_j$ to $x_i$ if and only if $b_{ij} \neq 0$.

Given a multivariate normal density, we can generate a Gaussian
belief network, and vice versa.
The unconditional means 
$\vec{m}$ are the same in both representations. 
Shachter and Kenley (1989) describe the general transformation 
from $\vecv$ and $\{b_{ij} \mid i < j \}$ of a given Gaussian belief 
network $G$ to the precision matrix $\matW$ of the 
normal distribution represented by $G$.  They use the following 
recursive 
formula in which $W(i)$ denotes the $i \times i$ upper left  
submatrix of $\matW$, $\vecb_i$ denotes the column vector 
$(b_{1,i}, \ldots, b_{i-1,i})$ and $\vecb'_i$ 
denotes the transposed vector $\vecb_i$ (i.e., the 
line vector $(b_{1,i}, \ldots, b_{i-1,i})$): 
\begin{equation} 
\label{eq:shachter} 
\matW(i+1) = \left( \begin{array}{cc} 
\matW(i) + \frac{\vecb_{i+1} \vecb'_{i+1}}{v_{i+1}} & 
-\frac{\vecb_{i+1}}{v_{i+1}}   \\ 
-\frac{\vecb'_{i+1}}{v_{i+1}} &  \frac{1}{v_{i+1}} 
\end{array} \right) 
\end{equation} 
for $i > 0$, and $\matW(1)= \frac{1}{v_1}$. 
\Eq{eq:shachter} plays a key role in this paper. 
 
\begin{figure}  
\begin{center} 
\leavevmode 
\includegraphics[width=2.0in]{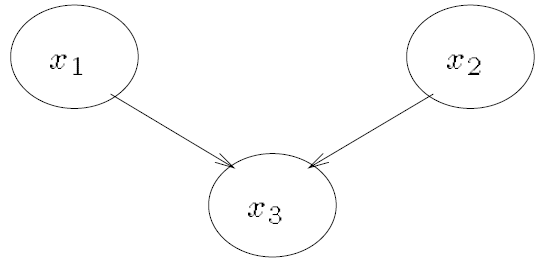}
\end{center}
\caption{A belief-network structure for three variables.} 
\label{fig:x}
\end{figure}

For example, suppose $x_1 = n(m_1, 1/v_1), 
x_2 = n(m_2, 1/v_2),$ and   
$x_3 = n(m_3 + b_{13} (x_1-m_1) +  
b_{23} (x_2-m_2), 1/v_3)$. 
The belief-network structure defined by these equations is shown in  
\Fig{fig:x}.  
The precision matrix is given by 
\begin{equation} 
\label{eq:covmat} 
W =  
\left( \begin{array}{ccc} 
\frac{1}{v_1} + \frac{b_{13}^2}{v_3} &  \frac{b_{13}b_{23}}{v_3} &   
  -\frac{b_{13}}{v_3} \\ 
\frac{b_{13}b_{23}}{v_3} & \frac{1}{v_2} + \frac{b_{23}^2}{v_3} &    
  -\frac{b_{23}}{v_3} \\ 
-\frac{b_{13}}{v_3} & -\frac{b_{23}}{v_3} & \frac{1}{v_3} 
\end{array} \right) 
\end{equation} 
 
The Gaussian-belief-network representation of a multivariate
normal distribution is better
suited to model elicitation and understanding than is
the standard representation
\cite{Shachter89b}.  To assess a Gaussian belief network, the user needs to specify (1) the unconditional mean of each variable $x_i$ ($m_i$),
(2) the relative importance of each parent $x_j$ in determining the values of its child $x_i$ ($b_{ij}$), and (3) a conditional variance for $x_i$ given that its parents are fixed ($v_i$).  \Eq{eq:shachter} then determines $\matW$.  In contrast, when assessing a normal distribution directly, one needs 
to guarantee that the assessed covariance matrix is  
positive-definite---a task done by altering in some {\em ad hoc} manner 
the correlations stated by the user.

\section{A Metric for Gaussian Belief Networks} \label{sec:bnm}

We are interested in computing a score for a 
Gaussian belief-network structure, given a set
of cases $D = \{\x_1,\ldots,\x_m\}$.  Each {\em case}
$\x_i$ is the observation of one or more variables
in $\x$.  We sometimes refer to $D$ as a {\em database}.
\Tab{tab:3nodec} is an example of a database for the three-node domain of the Gaussian belief network shown in \Fig{fig:x}.

\begin{table} 
\tiny
\caption{An complete database for the domain associated with the network shown in Figure~\protect\ref{fig:x}.}
\begin{center}
\begin{tabular}{|cccc|}
\hline
& \multicolumn{3}{c|}{Variable values for each case} \\ 
Case & $x_1$   & $x_2$   & $x_3$  \\ 
\hline
1  & -0.78 & -1.55 & 0.11 \\
2  & 0.18 & -3.04 & -2.35 \\
3  & 1.87 & 1.04 & 0.48 \\
4  & -0.42 & 0.27 & -0.68 \\
5  & 1.23 & 1.52 & 0.31 \\
6  & 0.51 & -0.22 & -0.60 \\
7  & 0.44 & -0.18 & 0.13 \\
8  & 0.57 & -1.82 & -2.76 \\
9  & 0.64 & 0.47 & 0.74 \\
10 & 1.05 & 0.15 & 0.20 \\
11 & 0.43 & 2.13 & 0.63 \\
12 & 0.16 & -0.94 & -1.96 \\
13 & 1.64 & 1.25 & 1.03 \\
14 & -0.52 & -2.18 & -2.31 \\
15 & -0.37 & -1.30 & -0.70 \\
16 & 1.35 & 0.87 & 0.23 \\
17 & 1.44 & -0.83 & -1.61 \\
18 & -0.55 & -1.33 & -1.67 \\
19 & 0.79 & -0.62 & -2.00 \\
20 & 0.53 & -0.93 & -2.92 \\
\hline
\end{tabular}
\end{center}
\label{tab:3nodec}
\end{table} 

Our scoring metrics are based on five assumptions, the
first of which is the following:

\begin{assumption} \label{ass:rand-samp} 
The database $D$ is a random sample from a multivariate 
normal distribution with unknown means $\m$  
and unknown precision matrix $\matW$. 
\end{assumption} 
 
Because every Gaussian belief network is equivalent to a multivariate 
normal distribution, Assumption~\ref{ass:rand-samp} is equivalent to 
stating that the database $D$ is a random sample from a Gaussian belief 
network with unknown parameters, $\vecv, B= \{b_{ij} | j < i \}, \m$. 
 
A Bayesian measure of the goodness of a network structure is its posterior probability given a database: 
\[
p(\Bs|D,\xi) = c \ p(\Bs|\xi) \ \rho(D|\Bs,\xi)
\]
where $c = 1/\rho(D|\xi) = 1/\sum_{\Bs} p(\Bs|\xi) \ \rho(D|\Bs,\xi)$
is a normalization constant.  For even small domains, however, there
are too many network structures to sum over in order to determine the
constant.  Therefore we use $p(\Bs|\xi) \ \rho(D|\Bs,\xi) = \rho(D,\Bs|\xi)$ as our score.

Also problematic is our use of the term $\Bs$ as an argument of
a probability.  In particular, $\Bs$ is a belief-network structure,
not an event.  Thus, we need a definition of an event $\eBs$ that
corresponds to structure $\Bs$ (the superscript ``$e$'' stands for 
$e$vent).  A natural definition for this event is that
$\eBs$ holds true iff the database is a random sample
from a {\em minimal} Gaussian belief network with structure $\Bs$
---that is, iff for all $j<i$, $b_{ij} \neq 0$ if and only if
there is an arc from $x_j$ to $x_i$ in $\Bs$.  For example the event $\eBs$ corresponding to the Gaussian belief network of \Fig{fig:x}, is the event $\{ b_{12}= 0, b_{13} \neq 0, b_{23} \neq 0\}$.

This definition has the following desirable property.
When two belief-network structures represent 
the same assertions of conditional independence, we say 
that they are {\em isomorphic}.  For example, in 
the three variable domain $\{x_1,x_2,x_3\}$, the 
network structures $x \rightarrow x_2\rightarrow x_3$ and 
$x_1 \leftarrow x_2\rightarrow x_3$ represent the same assertion: $x_1$ and
$x_3$ are independent given $x_2$.
Given the definition of $\eBs$, it can be shown 
that events $\eBsone$ and $\eBstwo$ are equivalent if and only if the 
structures $\Bsone$ and $\Bstwo$ are isomorphic.  That is, the relation of isomorphism induces an equivalence class on the set of events $\eBs$.   
We call this property {\em event equivalence}.

There is a problem with the definition, however.  In particular,
events corresponding to some non-isomorphic network
structures are not mutually exclusive.  
For example, in the
four-variable domain $\{x_1,x_2,x_3,x_4\}$, 
consider the structures 
$x_1 \Rightarrow B \Leftarrow x_4$ and 
$x_1 \Rightarrow B \Rightarrow x_4$, where $B$ is the subnetwork
structure $x_2 \rightarrow x_3$, and $x \Rightarrow B$ means
that there is an arc from $x$ to both variables in $B$.
The events corresponding to these structures
both include the situation where $x_1$ and $x_4$ are marginally independent.  Arbitrary overlaps between
events can make scores difficult to interpret
and use.  For example, the prediction of future events by averaging over multiple models cannot be justified.
In our case, however, we can repair the
definition of $\eBs$ so as to make non-equivalent events
mutually exclusive, without affecting our mathematical results or the intuitive understanding of events by the user.
In particular, all overlaps will be of measure zero
with respect to the events that create the overlap.
Thus, given a set of overlapping events, 
we simply exclude the intersection from all but one of the
events.  We note that this revised definition retains the
property of event equivalence.
 
\begin{proposition}[Event Equivalence]\label{prop:ee} 
Belief-network structures $\Bsone$ and $\Bstwo$ are isomorphic if and only if $\eBsone = \eBstwo$. 
\end{proposition}

Because the score for network structure $\Bs$ is $\rho(D,\eBs|\xi)$,
an immediate consequence of the property of event equivalence is
score equivalence.

\begin{proposition}[Score Equivalence]\label{prop:se} 
The scores of two isomorphic belief-network structures must be 
equal. 
\end{proposition} 

Given the property of event equivalence, we technically
should score each belief-network-structure equivalence class, rather 
than each belief-network structure.  Nonetheless, users find it 
intuitive to work with (i.e., construct and interpret) belief networks.  
Consequently, we continue our presentation 
in terms of belief networks, keeping Proposition~\ref{prop:se} in mind.

\subsection{Complete Gaussian Belief Networks} 
 
We first derive $\rho(D,\eBs|\xi)$, assuming $\Bs$ is the structure of a 
complete Gaussian belief network.   A {\em complete Gaussian belief 
network} is one with no missing edges.  
Applying the property of event equivalence, we know that the event 
associated with any complete belief network is the same; and we use 
$\eBsc$ to denote this event. 
 
To motivate the derivation, consider the following expansion of $\rho(D 
| \eBsc,\xi)$: 
\[
\rho( D | \eBsc, \xi)  =  
  \prod_{l=1}^m \rho(C_l | C_1, \ldots, C_{l-1}, \eBsc, \xi) =
\]
\[
\prod_{l=1}^m \int \  
  \rho(C_l | \vec{m}, \matW, \eBsc, \xi) \  \rho(\vec{m}, \matW | C_1, 
\ldots,    
    C_{l-1},\eBsc,\xi) \ d\vec{m} \ d\matW 
\]

\noindent Thus, we can derive the metric if we find a conjugate distribution for 
the parameters $\vec{m}$ and $\matW$ such that the integral above has a 
closed form solution.  

The next assumption leads to such a conjugate distribution.  If all variables in a case are observed, we say that the case is {\em complete.}  If all cases in a database are complete, we say that the database is {\em complete}.  

\begin{assumption} \label{ass:compl-c} 
All databases are complete.\footnote{SDLC present a survey of approximation methods
for handling missing data in the context of discrete variables.  Some of
these methods in modified form can be applied to Gaussian networks.} 
\end{assumption} 
 
Given this assumption, the following 
distribution is conjugate 
for multivariate-normal sampling. 

\begin{theorem}[DeGroot, p.178] \label{thm:DG} 
Suppose that $\xone,\ldots, \xl$ is a random sample from a 
multivariate normal distribution with an unknown value of the  
mean vector $\m$ and an unknown value of the precision matrix 
$\matW$.  Suppose that the prior joint distribution of $\m$ and $\matW$ 
is the normal-Wishart distribution: the conditional distribution of $\m$ 
given $\matW$ is $n(\vecmu_0, \nu \matW)$ such that $\nu > 0$, 
and the marginal distribution of $\matW$ is a Wishart distribution with 
$\alpha > n-1$ degrees of freedom and 
precision matrix $\matTau_0$, 
denoted by $w(\alpha,\matTau_0)$.  Then the posterior joint distribution 
of $\m$ and $\matW$ given $\vecxi$, $i=1,\ldots,l$, is 
as follows:  The conditional distribution of $\m$ given 
$\matW$ is a multivariate normal distribution 
with mean vector $\vecmu_l$ and a precision matrix 
$(\nu + l) \matW$, where 
\begin{equation} 
\label{eq:updatemeans} 
\Xl = \frac{1}{l} \sum_{i=1}^l \x_i, \;\;\;\; 
\vecmu_l = \frac{\nu \vecmu_0 + l \Xl}{\nu+l}. 
\end{equation} 
and the marginal of $\matW$ is $w(\alpha+l,\matTau_l)$, where 
$\matS_l$ and $\matTau_l$ are given by 
\begin{equation} \label{eq:int-scat}
\matS_l = \sum_{i=1}^l (\x_i - \Xl)(\x_i - \Xl)' 
\end{equation}
and 
\begin{equation} 
\label{eq:updateTau} 
\matTau_l =  
\matTau_0 + \matS_l + \frac{\nu l}{\nu+l}(\vecmu_0-\Xl)(\vecmu_0-\Xl)' 
\end{equation} 
\end{theorem} 

In this theorem, $\Xl$ and $\matS_l$ are the sample mean
and \textcolor{red}{scatter matrix} of the database, respectively.
Also,
an $n$ dimensional Wishart distribution with $\alpha$  
degrees of freedom and 
matrix $\matTau_0$ is given by 
\begin{equation} \label{eq:Wishart} 
\rho(W|\xi) = w(\alpha,\matTau_0)
 \equiv  c(n,\alpha) |\matTau_0|^{\alpha/2}
|\matW|^{(\alpha-n-1)/2} e^{-1/2\tr\{\matTau_0 \matW\}}
\end{equation}
where $\tr\{\matTau_0 \matW\}$ is the sum of the diagonal elements 
of $\matTau_0 \matW$ 
and 
\begin{displaymath} 
\label{eq-c-alpha-n} 
c(n,\alpha) = \left[ 2^{\alpha n /2} \pi^{n(n-1)/4} 
\prod_{i=1}^n \Gamma(\frac{\alpha+1-i}{2}) \right]^{-1} 
\end{displaymath} 
The parameters \textcolor{blue}{$\nu$,} $\alpha$, 
\textcolor{blue}{$\vecmu_0$,} and $\matTau_0$ are implicit
functions of the user's background knowledge $\xi$. 
\textcolor{blue}{The quantities $\nu$ and $\alpha$ can be thought of as 
the effective sample sizes of the normal and Wishart components of the prior, 
respectively.}
 
\comment{
From \Eq{eq:updatemeans}, we see that $\nu$ can be thought of as  
being an \textcolor{blue}{effective} sample size for $\vec{m}$---that is, the 
equivalent number of cases 
the user has seen, since he was ignorant about $\vec{m}$.  
When $l$ new cases are seen, the posterior mean 
is updated as a weighted average of the prior mean 
computed based on $\nu$ cases and the sample mean 
based on $l$ cases.  Furthermore, if $\vec{x_1},\ldots,\vec{x_{\alpha}}$ is a random sample  
of $n$-dimensional random vectors from a multivariate normal  
distribution for which the mean vector is {\boldmath 0} and  
the $n \times n$ matrix is $\matTau_0$,  
then $W = \sum_{i=1}^{\alpha} \vec{x_i} \vec{x_i}'$ has the  
Wishart distribution given in \Eq{eq:Wishart} (DeGroot, p.56).   
Thus, we may interpret $\alpha$ as the user's \textcolor{blue}{effective} sample size 
for the matrix $\matTau_0$.  Note that $\alpha$ must be  
at least the number of variables in the domain.   
We address the assessment of $\vecmu_0$ and $\matTau_0$ in \Sec{sec:prior-net-c}.}
 
Summarizing our discussion so far, we make the following assumption: 
 
\begin{assumption} \label{ass:wn} 
The prior distribution $\rho(\m,\matW|\eBsc,\xi)$  
is a normal-Wishart distribution as given in \Thm{thm:DG}. 
\end{assumption} 
 
\noindent From \Eq{eq:shachter}, this assumption fixes the distribution 
$\rho(\m,\vecv,\matB|\eBsc,\xi)$.  Nonetheless, we shall sometimes 
find it 
easier to specify the prior density in the space of $\matW$, rather then 
in the space of parameters describing a Gaussian belief network. 
 
If $\rho(\x | \m,\matW, \eBsc,\xi) = n(\m,\matW)$ 
and if $\rho(\m,\matW | \eBsc,\xi)$ is a normal-Wishart distribution as 
specified by Theorem~\ref{thm:DG}, then 
$\rho(\x | \eBsc,\xi)$, defined by 
\[ 
\rho(\x | \eBsc,\xi)= \int \rho(\x |\m,\matW, \eBsc,\xi) \ 
  \rho(\m,\matW,\eBsc,\xi) \ d\m \ d\matW
\] 
\noindent is an $n$ dimensional multivariate $t$ distribution with  
$\gamma = \alpha - n + 1$ degrees of freedom, 
location vector $\vecmu_0$, and a precision matrix 
$\matTau_0' = \frac{\nu \gamma}{\nu+1}\matTau_0^{-1}$.
\textcolor{blue}{
This result can be derived by first integrating over $\m$ using
Equation 6 on p.178 of DeGroot with sample size equal to one, and then
integrating over $\matW$
following an approach similar to that on pp.179--180 of DeGroot.
Also, using Equation 3 on p.180 of DeGroot,}
the $t$ distribution  
$\rho(\x | \eBsc,\xi)$ can be written in a less traditional 
form as follows: 
\begin{equation} \label{eq:t-dist} 
\rho(\x \mid \eBsc,\xi) 
 = 
  (2\pi)^{-n/2} (\frac{\nu}{\nu+1})^{n/2} 
  \frac{c(n,\alpha)}{c(n,\alpha+1)} |\matTau_0|^{\alpha/2}  
  |\matTau_1|^{-(\alpha+1)/2}
\end{equation}
\noindent where $\matTau_1$ is defined by \Eq{eq:updateTau} with $l=1$. 

Combining these facts with Theorem~\ref{thm:DG}, we know that 
$\rho(C_l | C_1, \ldots, C_{l-1},\eBsc,\xi)$ is a multivariate $t$  
distribution with parameters $\nu+l-1$, $\alpha+l-1$, $\vecmu_{l-1}$, 
and $\matTau_{l-1}$.  Consequently, we obtain  
\begin{eqnarray}  
\rho( D \mid \eBsc,\xi)  & = &
\prod_{l=1}^m \rho(C_l \mid C_1, \ldots, C_{l-1},\eBsc,\xi) 
\nonumber \\  
 & = & 
\prod_{l=1}^m \left( (2\pi)^{-n/2} (\frac{\nu+l-1}{\nu+l})^{n/2} 
\frac{c(n,\alpha+l-1)}{c(n,\alpha+l)} 
\frac{|\matTau_{l-1}|^{\frac{\alpha+l-1}{2}}} 
{|\matTau_l|^{\frac{\alpha+l}{2}}} \right) 
\nonumber \\  
& = & 
(2\pi)^{-nm/2} (\frac{\nu}{\nu+m})^{n/2} 
\frac{c(n,\alpha)}{c(n,\alpha+m)} 
|\matTau_0|^{\frac{\alpha}{2}} |\matTau_m|^{-\frac{\alpha+m}{2}} 
\label{eq:Gpunch} 
\end{eqnarray} 
\noindent Multiplying \Eq{eq:Gpunch} by the prior probability  $p(\eBsc|\xi)$
yields a metric for scoring \textcolor{blue}{$\eBsc$}.
 
\subsection{General Gaussian Belief Networks} 
 
We now consider an arbitrary Gaussian belief network $\Bs$.  To form a prior distribution for the parameters of $\Bs$, we make two 
additional assumptions: 
 
\begin{assumption}[Parameter Independence]
\label{ass:param-i-c} 
For every Gaussian belief network $\Bs$, 
$\rho(\vecv, \matB | \eBs,\xi) = \prod_{i=1}^n \rho(v_i, \vecbi | \eBs,\xi)$. 
\end{assumption} 
 
\noindent We note that this assumption  
is consistent 
with Assumption~\ref{ass:wn}, because 
if $\rho(W|\eBsc,\xi)$ is a Wishart distribution, then 
$\rho(\vecv, \matB | \eBsc,\xi)$, obtained from $\rho(W|\eBsc,\xi)$  
by using \Eq{eq:shachter} and the Jacobian $\dXdY{\matW}{\vecv \matB}$
of this transformation, is  
equal to $\prod_{i=1}^n \rho(v_i, \vecbi | \eBsc,\xi)$. 
The derivation of this claim is given  
in the Appendix (\Thm{thm:p-ind-w}). 
 
\begin{assumption}[Parameter Modularity] 
 \label{ass:param-m-c} 
If $x_i$ has the same parents in two  
Gaussian belief networks $\Bsone$ and $\Bstwo$, then 
$\rho(v_i, \vecbi|\eBsone,\xi) = \rho(v_i, \vecbi|\eBstwo,\xi)$.
\end{assumption} 

\Ass{ass:param-i-c} has been made in discrete contexts by
many researchers (e.g., CH, Buntine, SDLC, and HGC).  
\Ass{ass:param-m-c} has also been made by these
same researchers, but HGC were the first researchers to make the assumption
explicit and to emphasize its importance for generating
prior distributions.  Parameter modularity plays a similar
important role in the current development.  In particular,
this assumption, in conjunction with the
property of event equivalence and our previous assumptions
allows us to determine the joint prior distribution 
of the parameters $\m,\vecv,\matB$ associated with any 
Gaussian network $\Bs$ from the joint density
$\rho(\vec{m},\matW|\eBsc)$.

\begin{sloppypar}
To see this fact, first note
that, by the definition of the event $\eBs$, $\rho(\m|\vecv,\matB,\eBs,\xi)=\rho(\m|\vecv,\matB,\eBsc,\xi)$.  The latter distribution is
determined by $\rho(\m|W,\eBsc,\xi)$, which is given.
Second, from Assumption~\ref{ass:param-i-c}, we obtain
$\rho(\vecv,\matB | \eBs,\xi)$ by determining 
$\rho(v_i,\vecbi | \eBs,\xi)$ for each $i$.  By
Assumption~\ref{ass:param-m-c}, however, $\rho(v_i,\vecbi | \eBs,\xi)$
is equal to $\rho(v_i,\vecbi | \eBscp,\xi)$ for any complete
network structure $\Bscp$ where the parents of $x_i$ are the
same as are those in $\Bs$.  By event equivalence
and Assumption~\ref{ass:param-i-c}, we obtain $\rho(v_i,\vecbi | \eBscp,\xi)$ from the given density $\rho(W | \eBsc,\xi)$.
\end{sloppypar}
 
From Assumptions~\ref{ass:rand-samp} through~\ref{ass:param-m-c}, we  
derive $\rho(D | \eBs,\xi)$.   
To do so, we need the following theorem whose proof is provided  
in the Appendix. \textcolor{blue}{[Note: a derivation
from weaker assumptions is given in
D. Geiger and D. Heckerman, Parameter Priors for Directed Acyclic Graphical 
Models and the Characterization of Several Probability Distributions, 
{\em The Annals of Statistics}, 30: 1412-1440, Oct 2002.]} 
 
\begin{theorem} \label{thm:p-sep} 
If $\rho(\x | \m, \matW, D,\xi)$ is a  
multivariate normal distribution, and $\rho(\m | \matW, D, \eBs,\xi)$  
is a multivariate normal distribution with a precision matrix  
$\nu \matW$, $\nu> 0$, then 
$\rho(x_i | x_1,\ldots,x_{i-1}, \vecv, \matB, D, \eBs,\xi) = 
\rho(x_i | \Pi_i, v_i, \vec{b_i}, \Di, \eBsdp,\xi)$, 
where $\Bsdp$ is any network where $x_i$ has the same  
parents as in $\Bs$, and $\Di$ is the database $D$ restricted to the  
variables in $\{x_i\} \cup \Pi_i$. 
In particular, this claim holds for any complete  
Gaussian belief network $\Bsc = \Bsdp$ in which $\Pi_i$ and $x_i$  
appear before any other variables, and $\Pi_i$ appears before 
$x_i$. 
\end{theorem} 
 
Let $D_l= \{C_1,\ldots,C_{l-1}\}$ and  
$C_l$ be an instance of $x_1,\ldots,x_n$.  In the following derivation, 
we use $x_i$ and $\Pi_i$ to represent the instance of $x_i$ and $\Pi_i$ 
in the $l$th case.  \Thm{thm:p-sep} yields, 
\begin{eqnarray*}  
\rho(D | \vecv, \matB, \eBs,\xi) 
  & = & 
  \prod_{l=1}^m \prod_{i=1}^n  
    \rho(x_i | x_1, \ldots, x_{i-1}, \vecv, \matB, D_l, \eBs,\xi)   
    \nonumber  \\ 
  & = & \prod_{l=1}^m \prod_{i=1}^n  
    \frac{\rho(x_i , \Pi_i| v_i, \vec{b_i}, \Dil, \eBs,\xi)}{ 
          \rho(\Pi_i | v_i, \vec{b_i}, \Dil, \eBs,\xi)} 
    \label{eq:p-fact1x} 
\end{eqnarray*} 
and 
\begin{displaymath} \label{eq:p-fact2x} 
\rho(\Pi_i | v_i, \vec{b_i}, \Dil, \eBs,\xi) =  
  \rho(\Pi_i | v_i, \vec{b_i}, D^{\Pi_i}_l, \eBs,\xi) 
\end{displaymath} 
By combining these equations, we obtain  
the following {\em likelihood 
separability property:} 
\begin{equation}  \label{eq:p-fact3x} 
\rho(D | \vecv, \matB, \eBs,\xi) = \prod_{i=1}^n  
  \frac{\rho(\Di|v_i, \vec{b_i}, \eBs,\xi)}{ 
        \rho(D^{\Pi_i}|v_i, \vec{b_i}, \eBs,\xi)} 
\end{equation} 
 
By Bayes rule, $\rho(\vecv, \matB | D, \eBs,\xi)$ is proportional 
to $\rho(D | \vecv,\matB,\eBs,\xi) \rho(\vecv,\matB | \eBs,\xi)$.  Thus,  
because $\rho(D | \vecv,\matB,\eBs,\xi)$ factors as shown by 
\Eq{eq:p-fact3x}, and 
$\rho(\vecv,\matB | \eBs,\xi)$ factors as given by 
Assumption~\ref{ass:param-i-c}, we obtain the following 
{\em posterior parameter independence} property: 
\[ 
\rho(\vecv, \matB | D, \eBs,\xi) =  
  \prod_{i=1}^n \rho(v_i, \vecbi | \Di, \eBs,\xi) 
\] 
In a similar manner, whenever $x_i$ has the same parents  
in two Gaussian belief networks $\Bs$ and $\Bsdp$, 
by using \Eq{eq:p-fact3x}  
where $\eBs$ in the right hand side  
is replaced by $\eBsdp$ and using Assumption~\ref{ass:param-m-c}, 
we obtain the  
{\em posterior parameter modularity} property: 
\[ 
\rho(v_i, \vecbi | \Di, \eBs,\xi) = \rho(v_i, \vecbi | \Di, \eBsdp,\xi) 
\] 
 
Now, we have 
\begin{eqnarray}
\rho( D | \eBs,\xi) & = &
\prod_{l=1}^m \rho(C_l | D_l,\eBs,\xi), \label{eq:G-productone} \\
\rho(C_l | D_l, \eBs,\xi) & = &
\prod_{i=1}^n \rho(x_i | x_1, \ldots,x_{i-1},D_l,\eBs,\xi) \nonumber \\
\rho(x_i | x_1, \ldots,x_{i-1},D_l,\eBs,\xi) & = & \int 
  \rho(x_i | x_1, \ldots,x_{i-1},D_l,\vecv,\matB,\eBs,\xi) 
 \rho(\vecv,\matB \mid D_l, \eBs,\xi)
    d\vecv\matB \label{eq:G-producthree}
\end{eqnarray} 
By applying \Thm{thm:p-sep} to the first term  
of the right-hand-side of \Eq{eq:G-producthree}, and 
posterior parameter independence and posterior 
parameter modularity to the second term, 
we obtain 
\begin{eqnarray*} 
\rho(x_i | x_1, \ldots,x_{i-1},D_l,\eBs,\xi) & = &   
  \int 
  \rho(x_i | \Pi_i, v_i, \vec{b_i}, \Dil, \eBsc,\xi) 
  \ \rho(v_i, \vec{b_i}| \Dil, \eBsc,\xi)
  \ dv_i \vecbi \\
& = & \rho(x_i|\Pi_i,\Dil,\eBsc,\xi) 
\end{eqnarray*} 
Therefore, 
\begin{equation} 
\rho(C_l | D_l, \eBs,\xi) = \prod_{i=1}^n 
  \frac{\rho(x_i,\Pi_i|\Dil,\eBsc,\xi)}{\rho(\Pi_i|\Dil,\eBsc,\xi)} \\    
  \label{eq:G-product4} 
\end{equation} 
Furthermore, because $\rho(\Pi_i|\Dil,\eBsc,\xi)$ is a  
multivariate $t$ distribution, we know that 
\[ 
\rho(\Pi_i|\Dil,\eBsc,\xi) = \rho(\Pi_i|\DiPi,\eBsc,\xi) 
\] 
(DeGroot, p.60). 
Thus, combining \Eqs{eq:G-productone} and \ref{eq:G-product4}, we have 
\begin{equation} \label{eq:G3-score} 
\rho(D | \eBs,\xi) = \prod_{i=1}^n 
  \frac{\rho(\Di | \eBsc,\xi)}{\rho(D^{\Pi_i} | \eBsc,\xi)} 
\end{equation} 
where each term in \ref{eq:G3-score} is of the form  
given in \Eq{eq:Gpunch}. 
Multiplying \Eq{eq:G3-score} by $p(\eBs|\xi)$, we obtain
a metric for an arbitrary Gaussian belief network $\Bs$. 
\textcolor{blue}{(This development is incomplete, as it requires
a recipe for deriving the parameters of the prior for subsets
of the domain variables from the prior for all domain variables.
The recipe implicit in an example given in the original version---deleted
in this version---is incorrect. For a correction, see 
the \href{https://arxiv.org/abs/2105.03248}{2021 update} of 
D. Geiger and D. Heckerman, Parameter Priors for Directed Acyclic Graphical 
Models and the Characterization of Several Probability Distributions, 
{\em The Annals of Statistics}, 30: 1412-1440, Oct 2002.)}
We call this metric \BGe\, which stands for $B$ayesian metric
for $G$aussian networks having score $e$quivalence.


\subsection{Score Equivalence}

In making the assumptions of parameter independence and parameter modularity, we have---in effect---specified the prior densities for the multinomial parameters in terms of the structure of a belief network. Consequently, there is the possibility that this specification violates the property of score equivalence.  The following theorem, however, demonstrates that our specification implies score equivalence.

\begin{theorem}[Score Equivalence]
If $\Bsone$ and $\Bstwo$ are isomorphic belief-network structures, then
$\rho(D|\eBsone, \xi )$ and $\rho(D|\eBstwo, \xi )$ as computed by
\Eq{eq:G3-score} are equal.
\end{theorem}

\noindent {\bf Proof:} In Heckerman et al. (1994, Theorem 10), we show that a belief network structure can be transformed into an isomorphic structure by a series of arc reversals, such that, whenever an arc from $x_i$ to $x_j$ is reversed, $\Pi_i = \Pi_j \setminus \{x_i\}$.  Thus, our claim follows if we can prove it for the case where $\Bsone$ and $\Bstwo$ differ by a single arc reversal with this restriction.

So, let $\Bsone$ and $\Bstwo$ be two isomorphic network structures that differ only in the direction of the arc between $x_i$ and $x_j$ (say $x_i \rightarrow x_j$ in $\Bsone$).  Let $R$ be the parents of $x_i$ in $\Bsone$.  By the cited theorem, $R \cup \{x_i\}$ is the parents of $x_j$ in $\Bsone$, $R$ is the parents of $x_j$ in $\Bstwo$, and 
$R \cup \{x_j\}$ is the parents of $x_i$ in $\Bstwo$.  Because the two structures differ only in the reversal of a single arc, the only terms in the product of \Eq{eq:G3-score} that can differ are those involving $x_i$ and $x_j$.
For $\Bsone$, these terms are
\[
\frac{\rho(D^{x_i R}|\eBsc,\xi)}{\rho(D^{R}|\eBsc,\xi)}
  \frac{\rho(D^{x_i x_j R}|\eBsc,\xi)}{\rho(D^{x_i R}|\eBsc,\xi)}
     = \frac{\rho(D^{x_i x_j R}|\eBsc,\xi)}{\rho(D^{R}|\eBsc,\xi)}
\]
whereas for $\Bstwo$, they are
\[
\frac{\rho(D^{x_j R}|\eBsc,\xi)}{\rho(D^{R}|\eBsc,\xi)}
  \frac{\rho(D^{x_i x_j R}|\eBsc,\xi)}{\rho(D^{x_j R}|\eBsc,\xi)}
     = \frac{\rho(D^{x_i x_j R}|\eBsc,\xi)}{\rho(D^{R}|\eBsc,\xi)}
\]
Thus, $\rho(D|\eBsone,\xi)=\rho(D|\eBstwo,\xi)$. \qed

\subsection{Encoding Prior Knowledge: The Prior Gaussian Belief Network} \label{sec:prior-net-c} 

From the previous discussion, we see that there are three components
of a user's prior knowledge that are relevant to learning Gaussian
networks: (1) the prior probabilities $p(\eBs|\xi)$, (2) the 
\textcolor{blue}{effective} sample
sizes $\alpha$ and $\nu$, and (3) the parameters $\vecmu_0$ and $\matTau_0$.  The assessment of the prior probabilities $p(\eBs|\xi)$
is straightforward.  Buntine and HGC, for example, describe methods that facilitate these assessments.  In addition, a user can assess the 
\textcolor{blue}{effective} sample sizes directly.  In this section, we concentrate on the assessment of $\vecmu_0$ and $\matTau_0$.

\textcolor{blue}{Using (1) our previous observation that
$p(\vecx|\eBsc,\xi)$ is a multivariate $t$ distribution, and (2) Equation 11
on p.61 of DeGroot with $\alpha > n+1$, we obtain
\begin{equation} \label{eq:t1}
{\rm E}(\vecx|\eBsc,\xi) = \vecmu_0 \ \ \ \ \ \ 
{\rm Cov}(\vecx|\eBsc,\xi) = \frac{\nu+1}{\nu} \ \frac{1}{\alpha-n-1} \ T_0
\end{equation}
Thus, a person can assess 
a Gaussian belief network 
for E$(\vecx|\eBsc,\xi)$ and Cov$(\vecx|\eBsc,\xi)$, and
then compute $\vecmu_0$ and $\matTau_0$ using Equations~\ref{eq:t1}.
We call this belief network a {\em prior belief network}.
}

\section{Metrics for Gaussian Causal Networks} \label{sec:cnm} 
 
People often have knowledge about the causal  
relationships among variables 
in addition to knowledge about conditional independence.  Such causal 
knowledge is stronger than is  
conditional-independence knowledge, because 
it allows us to derive beliefs about a domain after we intervene.
Causal networks, described---for example---by 
Spirtes et al. (1993)\nocite{Spirtes93},
Pearl and Verma (1991)\nocite{Pearl91}, and 
Heckerman and Shachter (1994)\nocite{HS94uai} 
represent such causal relationships among variables.  In particular, a 
causal network for $U$ is a belief network for $U$, wherein it is 
asserted that each nonroot node $x$ is caused by its parents. 
The precise meaning of cause and effect is not important for our 
discussion.  The interested reader should consult the previous 
references.   
 
The event $\eCs$ is the same as that for a belief-network structure,
except that we also include in the event the assertion that each nonroot node is caused by its parents.  Thus, in contrast to the
case for belief networks, it is not appropriate to require the properties of event equivalence or score equivalence.  
For example, consider a domain containing two variables $x$ and $y$.  Both the causal network $\Csone$ where $x$ points to $y$ and the causal network $\Cstwo$ where $y$ points to $x$ represent the assertion that $x$ and $y$ are dependent.  The network $\Csone$, however, in addition represents the assertion that $x$ causes $y$, whereas the network $\Cstwo$ represents the assertion that $y$ causes $x$.  Thus, the events $\eCsone$ are $\eCstwo$ are not equal.  Indeed, it is reasonable to assume that these events---and the events associated with any two different causal-network structures---are mutually exclusive.

In principle, then, a user may assign a (possibly different) prior distribution 
to the parameters $\vec{m}$, $\vecv$, and $\matB$ to every complete 
Gaussian causal network, constrained only by the assumption
of parameter modularity.  The prior distributions for parameters of
incomplete networks would then be determined by parameter modularity.  We call this
general metric \BG, as it is a superset of the \BGe\ metric.
For practical reasons, 
however, the assessment process should be constrained.  One
alternative is to use the \BGe\ metric.  A more
general alternative is to continue
to use the prior network to compute $\vecmu_0$ and $\matTau_0$,
but to allow \textcolor{blue}{effective} sample size to vary for different
variables and different parent sets of each variable.
We call this metric the \BGp\
metric, where ``p'' stands for $p$rior network.

\section{Summary and Future Work}

We have described metrics for learning belief networks and causal networks from a combination of user knowledge and statistical data
for domains containing only continuous variables.
An important contribution has been our elucidation of the property of event equivalence and the assumption of
parameter modularity.  
We have shown that these 
properties, when combined, allow a statistician to compute a reasonable prior 
distribution for the parameters of 
any Gaussian belief network, given a single prior Gaussian belief 
network provided by a user.

A legitimate concern with our approach is that the multivariate
model is too restrictive.  In practice, when this model
is inappropriate, statisticians will typically turn to a more
general model where each continuous variable conditioned on its parents is assumed to be a mixture of multivariate normal distributions.
In Geiger and Heckerman (1994)\nocite{GH94tr}, we derive
metrics for domains containing both discrete and continuous
variables, subject to the restriction that a domain
can be decomposed into disjoint sets of continuous variables 
where each such set is conditioned by a set of discrete variables.
We note that this work, when combined with approximation
methods that handle missing data, provides a method for
learning with multivariate mixtures.

In the discrete case, a complete network has one parameter  
for each instance of $\x$.  Consequently, it is easy to  
overfit such a structure with data; and the metrics developed for
discrete domains
provide a means by which we  
can avoid such overfitting.   
In the continuous case, a complete network has only $n + n(n-1)/2$  
parameters.  Thus, it is possible that the errors introduced  
by our methods, arising from heuristic search  
in an exponential space to find one or a handful of structures with high  
scores outweigh the benefits associated  
with decreasing the degree of overfitting.  We leave 
this concern for future experimentation.

\section*{Acknowledgments}

We thank Wray Buntine and anonymous reviewers for useful suggestions.

\bibliographystyle{apalike} 

\section*{Appendix} 
 
\begin{theorem} \label{thm:jac-w} 
The Jacobian $J$ for the change of variables from $W$ to $\{\vecv, 
\matB\}$ is given by 
\begin{equation}  \label{eq:jac-w} 
J = \dXdY{\matW}{\vecv \matB} =  
  \prod_{i=1}^n v_i^{-(i+1)} 
\end{equation} 
\end{theorem} 
 
\noindent {\bf Proof:} Let $J(i)$ denote the Jacobian for the first $i$ 
variables in $\matW$.  Then $J(i)$ has the following matrix form: 
\begin{equation} 
\label{eq:G-jacmatrix} 
\left( \begin{array}{ccc} 
J(i-1)  & 0 & 0  \\ 
0       & -\frac{1}{v_i} I_{i-1,i-1} & 0  \\ 
0       &  0          & -\frac{1}{v_i^2} 
\end{array} \right) 
\end{equation} 
where $I_{k,k}$ is the identity matrix of size $k \times k$. 
Thus, the absolute value of $J(i)$ is given by, 
\begin{equation} 
\label{eq:G-jacobian} 
|J(i)| = \frac{1}{v_i^{i+1}} \cdot |J(i-1)|  
\end{equation} 
which gives \Eq{eq:jac-w}. \qed

\begin{theorem} \label{thm:p-ind-w} 
If $\rho(\matW|\xi)$ has an n-dimensional Wishart distribution, then 
\[ 
\rho(\vecv, \matB|\xi) = \prod_{i=1}^n \rho(v_i, \vecbi|\xi) 
\] 
\end{theorem} 
 
\noindent {\bf Proof:} By assumption, we have 
\begin{equation} \label{eq:p-ind-w1} 
\rho(\matW|\xi) =  
  c \ |\matW|^{(\alpha-n-1)/2} e^{-1/2\tr\{\matTau_0 \matW\}}  
\end{equation} 
Thus, we must express \Eq{eq:p-ind-w1} in terms of $\{\vecv, \matB\}$, 
multiply by the Jacobian given by \Thm{thm:jac-w}, and show that the 
resulting function factors as a function of $i$.  From \Eq{eq:shachter}, 
we get 
\[ 
|\matW(i)| = \frac{1}{v_i}|\matW(i-1)| = \prod_{i=1}^n v_i^{-1} 
\] 
so that the determinant in \Eq{eq:p-ind-w1} factors as a function of 
$i$.  Also, \Eq{eq:shachter} implies (by induction) that each element 
$w_{ij}$ in $\matW$ is a sum of terms each being a function of $\vecbi$ 
and $v_i$. Consequently, the exponent in \Eq{eq:p-ind-w1} factors as a 
function of $i$. \qed

\begin{oldtheorem}{thm:p-sep} 
If $\rho(\x | \m, \matW, D, \eBs,\xi)$ is a multivariate  
normal distribution, and $\rho(\m | \matW, D, \eBs,\xi)$ is a multivariate  
normal distribution with precision matrix $\nu \matW$, $\nu > 0$, then 
$\rho(x_i | x_1,\ldots,x_{i-1}, \vecv, \matB, D, \eBs,\xi) = 
\rho(x_i | \Pi_i, v_i, \vec{b_i}, \Di, \eBsdp,\xi)$ 
where $\Bsdp$ is any network where $x_i$ has the same parents as  
in $\Bs$, and $\Di$ is the database $D$ restricted to the  
variables in $\{x_i\} \cup \Pi_i$. 
\end{oldtheorem} 

\noindent {\bf Proof:}   
Using
\[\rho( \x| \matW, D, \eBs,\xi) =  
  \int \rho( \x| \m, \matW, D, \eBs,\xi) \  
  \rho( \m | \matW, D, \eBs,\xi) \ d\m 
\] 
\noindent and Assumptions~\ref{ass:rand-samp} and \ref{ass:wn}, we obtain 
\begin{equation} \label{eq:p-sep2} 
\rho(\x | \matW, D, \eBs,\xi) 
 = 
  c \  |\matW|^{1/2} \cdot 
  e^{-\frac{1}{2}\frac{\nu}{\nu+1} 
    \sum_{i,j=1}^n (x_i-\mu_{Di})(x_j-\mu_{Dj})w_{ij} }
\end{equation}
where $\vecmuD$ is the posterior mean after seeing $D$, 
given by \Eq{eq:updatemeans} of \Thm{thm:DG}. 
 
The marginal distribution $\rho(x_1,\ldots,x_i|\xi)$ 
of a normal distribution $n(\m,\matW)$ 
is a normal distribution $n(\m_i, \matW_i)$, where $\m_i$ 
and $\matW_i$ are the terms in $\m$ and $\matW$ that 
correspond to $x_1,\ldots,x_i$.  Thus, using 
$|\matW|= \prod_{i=1}^n v_i^{-1}$, \Eq{eq:p-sep2} becomes 
\begin{equation} \label{eq:p-sep3} 
\rho(x_1, \ldots, x_i | \matW, D, \eBs,\xi) 
 = 
  c \ |\matW_i|^{1/2} \cdot 
  e^{-\frac{1}{2}\frac{\nu}{\nu+1} 
    \sum_{j,k=1}^i(x_j-\mu_{jD})(x_k-\mu_{kD}) w_{jk} } \nonumber
\end{equation}
By expressing $\matW$ in terms of $\vecv$ and $\matB$ 
using \Eq{eq:shachter}, we obtain 
\begin{equation} \label{eq:p-sep4} 
\frac{\rho(x_1,\ldots, x_{i} | \vecv, \matB, D, \eBs,\xi)}{ 
      \rho(x_1,\ldots, x_{i-1} | \vecv, \matB, D, \eBs,\xi)} =  
  c \cdot v_i^{-1/2} \cdot 
  e^{-\frac{1}{2}\frac{\nu}{\nu+1}A} 
\end{equation} 
where 
\begin{equation} \label{eq:p-sep5} 
A = \tr \left[ (\x-\vecmuD)_i (\x - \vecmuD)'_i  
  \left( 
    \begin{array}{cc} 
      \frac{\vecb_{i} \vecb'_{i}}{v_{i}} & -\frac{\vecb_i}{v_i}   \\ 
     -\frac{\vecb'_{i}}{v_{i}}           &  v_{i} 
    \end{array}  
  \right) 
\right] 
\end{equation} 
where $(\x-\vecmuD)_i$ is the column vector of the $i$  
elements of $(\x-\vecmuD)$ that correspond to $x_1,\ldots,x_i$.   
Starting with any network $\Bsdp$,  
such that the parents of $x_i$ are the same as in $\Bs$, we 
obtain exactly \Eqs{eq:p-sep4} and \ref{eq:p-sep5}. 
Furthermore, because $\vecmuD$ depends only on $\Di$, 
the theorem is established.  \qed 
 
\end{document}